# A Model for Reasoning with Uncertain Rules in Event Composition Systems

**Segev Wasserkrug**\*
IBM Research,
Haifa, Israel

**Avigdor Gal**
Technion
Haifa, Israel

**Opher Etzion**
IBM Research,
Haifa, Israel

## Abstract

In recent years, there has been an increased need for the use of active systems - systems required to act automatically based on *events*, or changes in the environment. Such systems span many areas, from active databases to applications that drive the core business processes of today's enterprises. However, in many cases, the events to which the system must respond are not generated by monitoring tools, but must be inferred from other events based on complex temporal predicates. In addition, in many applications, such inference is inherently uncertain. In this paper, we introduce a formal framework for knowledge representation and reasoning enabling such event inference. Based on probability theory, we define the representation of the associated uncertainty. In addition, we formally define the probability space, and show how the relevant probabilities can be calculated by dynamically constructing a Bayesian network. To the best of our knowledge, this is the first work that enables taking such uncertainty into account in the context of active systems. Therefore, our contribution is twofold: We formally define the representation and semantics of event composition for probabilistic settings, and show how to apply these extensions to the quantification of the occurrence probability of events. These results enable any active system to handle such uncertainty.

## 1 INTRODUCTION

In recent years, there has been growing need for the use of active systems – that is, systems required to act automatically based on events. While the earliest active systems were in the realm of databases, a current major need for such active functionality is in the category of *Business Activity Monitoring* (*BAM*) and *Business Process Management* (*BPM*). These are systems that monitor, streamline and optimize the processes and activities which are at the heart of every business enterprise, and they comprise one of the largest of today's emerging markets. Concurrent with the proliferation of such applications, the events to which active systems must respond have expanded from IT and application-level events to business-level events.

In order for such an active system to react automatically to all events of interest, it must be able to recognize when such events occur. However, in many cases, these events are not generated by monitoring tools, and so must be inferred from other events. In an effort to make such reasoning about events possible, several event composition systems and prototypes have been defined. These systems include both a specification language designed to represent information regarding relations between events, and a runtime engine for carrying out the actual event inference. However, all existing mechanisms allow reasoning about deterministic relations among events only by, *e.g.*, representing this knowledge as a set of deterministic complex temporal predicates. As a result, the uncertainty inherent in many real-life examples can neither be represented nor reasoned about in existing mechanisms. For example, consider a banking system where (a) events regarding the purchase and sale of stocks are signaled by outside sources, but (b) the system is required to respond automatically to illegal stock trading occurrences. In such a case, the system cannot be certain whether illegal stock trading actually took place. Rather, the best that can be achieved is the quantification of some measure of likelihood regarding the occurrence of illegal trading events.

In general, there are two main types of uncertainty relevant to event composition systems. The first, which we term *imprecise information uncertainty*. This uncertainty is caused by imprecise information regarding events signalled by event sources. Possible causes for such uncertainty include faulty or imprecise sensors. The second, termed *uncertain relations between events*, involves the non-determinism inherent in the *relations* between events, as in the relation between the signalled and inferred events in the stock trading example above. Active systems must be able both to represent and to quantify such uncertainty.

---

\* Technion, Haifa, Israel

In this work, we formally define a class of specification languages that make it possible to represent both types of uncertainty, using probability as the uncertainty representation mechanism. In addition, we define the general principles underlying a reasoning mechanism for inferring the occurrence probabilities of events. To illustrate these principles, a specific inference algorithm for a specific language is detailed. To the best of our knowledge, this is the first work in the context of event composition languages that enables representing uncertainty, and reasoning about it, in a general and formal manner.

The rest of this article is organized as follows: Section 2 reviews existing research in the realms of event composition languages and probabilistic reasoning. Section 3 formally defines the notion of events, and defines the representation of the information relevant to each event. Section 4 specifies the representation and semantics of uncertain relations between events. Section 5 describes the actual calculation of occurrence probabilities for events. We close with Section 6, which summarizes the article, briefly describes how the formal framework can be used to create an uncertainty-handling component in an active application, and discusses possible future work. Throughout this paper, we use examples of an active system in the area of stock trading for illustration and clarification.

## 2 RELATED WORK

This section describes two classes of related research. Section 2.1 describes existing event composition systems and prototypes. Section 2.2 discusses mechanisms for reasoning about probability.

### 2.1 EVENT COMPOSITION

Event composition is supported by systems from various domains. Some are designed for active databases (*e.g.*, ODE (Gehani et al., 1992)) while others are general-purpose event composition languages (*e.g.*, the Situation Manager Rule Language (Adi and Etzion, 2004)). All existing languages enable deterministic inference of events, based on a set of rules. Each rule describes a complex temporal predicate or function, based on which inference is carried out.

A major shortcoming of all existing specification languages is that they can reason only about deterministic knowledge regarding the relations between events. In addition, none of the existing systems are designed to handle uncertainty in a general and formal manner.

### 2.2 MECHANISMS FOR PROBABILISTIC REASONING

The most common approach for quantifying probabilities are Bayesian (or belief) networks (Pearl, 1988). However, Bayesian networks are only adequate for representing propositional probabilistic relationships between entities. In addition, standard Bayesian networks cannot explicitly model temporal relationships. To overcome these limitations, several extensions to Bayesian networks have been defined, including Dynamic Belief Networks (Kjaerulff, 1992), Time Nets (Kanazawa, 1991), Modifiable Temporal Belief Networks (Constantin and Gregory, 1996) and Temporal Nodes Bayesian Networks (Gustavo and Luis, 1999). Although these extensions are more expressive than classical Bayesian networks, they nonetheless lack the expressive power of first-order logic. In addition, some of these extensions allow more expressive power at the expense of efficient calculation.

Another formal approach to reasoning about probabilities involves probabilistic logics (*e.g.*, (Bacchus, 1990) and (Halpern, 1990)). These enable assigning probabilities to statements in first-order logic, as well as inferring new statements based on some axiomatic system. However, they are less suitable as mechanisms for the calculation of probabilities in a given probability space.

A third paradigm for dealing with uncertainty using probabilities is the KBMC paradigm (Breese et al., 1994). This approach combines the representational strength of probabilistic logics with the computational advantages of Bayesian networks. In this paradigm, separate models exist for probabilistic knowledge specification and probabilistic inference – *i.e.*, probabilistic knowledge is represented in some knowledge model (usually a specific probabilistic logic), and whenever an inference is carried out, an inference model would be constructed based on this knowledge. In this work, we have chosen an approach very similar to this paradigm: Knowledge is represented as probabilistic rules (see Section 4), while probability calculation is carried out by constructing a Bayesian network based inference model (see Section 5).

## 3 EVENT MODEL

We distinguish between actual events, and the information held by an event composition system about events of interest. Section 3.1 defines the notion of an event in event composition systems, and Section 3.2 describes the information relevant to each event and its representation. In the rest of this paper, we will refer to actual events by lowercase letters, *e.g.*, $e$ and $\varepsilon$, while uppercase letters (*e.g.*, $E$) will be used to refer to the information the system has about an event.

### 3.1 EVENTS

In the context of active systems, an *event* is defined as an occurrence that is significant (falls within a domain of interest to the system), instantaneous (takes place at a specific point in time), and atomic (either occurs or not). Examples of events include notifications of changes in a stock price, failure of IT components, and a person entering or leaving a certain geographical area.

A variety of data can be associated with the occurrence of an event. Some data types are shared by all events, *e.g.*, the point in time at which the event occurred. Other data types are specific to certain event types only, *e.g.*, the name of a stock on the stock market. Two events that have exactly the same associated data types are said to be of the same *event type*. For example, consider events that quote the change in price of a specific stock. With each instance of this type of event, the relevant information consists of the name of the stock for which the price was quoted, and the new price. Therefore, all such events can be said to belong to the event type *stockQuote*.

In the context of event composition systems, a distinction is made between two types of events, namely *explicit events* and *inferred events*. Explicit events are those events for which an event notification is signaled by an **external** event source. Inferred events are inferred from explicit events using a set of rules.

### 3.2 REPRESENTING INFORMATION ABOUT EVENTS

We represent the information a composite event system holds about each event instance with a data structure we term *Event Instance Data* (*EID*). *EID* incorporates all relevant data about an event, including its type, time of occurrence, etc. In event composition systems with no uncertainty, each event can be represented by a single tuple of values $Val = \langle val_1, \ldots, val_n \rangle$ (one value for each type of data associated with the event). In our case, to capture the uncertainty associated with an event instance, the *EID* of each event instance is a Random Variable (*RV*). The possible values of *EID* are taken from the domain $V = \{notOccurred\} \cup V'$, where $V'$ is a set of tuples of the form $\langle val_1, \ldots, val_n \rangle$.

The semantics of a value of $E$ (encoded as an *EID*), representing the information the system has about event $e$ are as follows: The probability that the value of $E$ belongs to a subset $S \subseteq V \setminus \{notOccurred\}$ is the probability that event $e$ has occurred, and that the value of its attributes is some tuple of the form $\langle val_1, \ldots, val_n \rangle$, where $\{\langle val_1, \ldots, val_n \rangle\} \in S$. Similarly, the probability associated with the value $\{notOccurred\}$ is the probability that the event did not occur. As an example, consider an event that quotes a price of $100 for a share of IBM stock at time 10:45. Say that the system considers the following possible: The event did not occur at all; the event occurred at time 10:34 and the price of the stock was $105; and the event occurred at time 10:45 and the price was $100. In addition, say that the system considers the probabilities of these possibilities to be 0.3, 0.3 and 0.4, respectively. Then, the event can be represented by an RV $E$ whose possible values are $\{notOccurred\}, \{10:34, IBM, 105\}$ and $\{10:45, IBM, 100\}$. Also, $\Pr(E = \{notOccurred\})$ and $\Pr\{E = \{10:34, IBM, 105\}\}$ are both 0.3 and $\Pr\{E \in \{\{10:45, IBM, 100\}, \{10:34, IBM, 105\}\}\} = 0.7$.

An additional concept, relevant in the context of event composition systems, is that of *Event History* (*EH*). An event history $EH_{t_1}^{t_2}$ is the set of all events (of interest to the system), as well as their associated data, whose occurrence time falls between $t_1$ and $t_2$. For example, consider the following events: an event $e_1$, at time 10:30, quoting the value of an IBM share as $100; event $e_2$, at time 10:45, quoting the value of an IBM share as $105; event $e_3$ at time 11:00, quoting the value of an IBM share as $103. Using the notation described above, the events $e_1, e_2, e_3$ can be described by the tuples $\{10:30, IBM, 100\}$, $\{10:45, IBM, 105\}$, and $\{11:00, IBM, 103\}$. Examples of event histories defined on these events are the following: $EH_{10:30}^{10:45} = \{e_1, e_2\}$, $EH_{10:45}^{11:00} = \{e_2, e_3\}$, $EH_{10:30}^{10:35} = \{e_1\}$ and $EH_{10:30}^{11:00} = \{e_1, e_2, e_3\}$. Note that there does not exist an event history that consists of both $e_1$ and $e_3$, and that does not include $e_2$.

The actual event history is not necessarily equivalent to the information regarding the event history possessed by the system. We will therefore denote the latter by *system event history*.

In what follows, *e.attributeName* denotes the value of a specific attribute of a specific event $e$. For example, $e_1.occT$ refers to the occurrence time of event $e_1$. In addition, the type of event $e$ is denoted by $e \in type$.

## 4 RULE MODEL

Rules represent information on event relationships. Each rule serves as a template and can be applied at a given time $t$ to event histories that are known at that time. For simplicity sake, we assume that the result of applying a rule to a specific event history may serve for the inference of, at most, a single event.[1] A rule $r$ is given in the form $\langle sel_r^n, pattern_r^n, eventType_r, mappingExpressions_r, prob_r \rangle$ where:

$sel_r^n$ is a deterministic predicate returning a subset of an event history of size less than or equal to $n$ (for some integer $n$). If the returned subset has strictly less then $n$ elements, no further evaluation of the rule is carried out. A possible selection expression is "the first two events of type *stockQuote*." Therefore, for the event history $e_1, e_2, e_3$, if only $e_1$ is of type *stockQuote* the rule is not triggered. However, if both $e_1$ and $e_3$ are of type *stockQuote*, then the subset $\{e_1, e_3\}$ is selected, and evaluation of the rule continues.

$pattern_r^n$ is a (possibly complex temporal) predicate of arity $n$ over event instances (note that this is the same $n$ appearing in $sel_r^n$). This predicate is applied to the $n$ event instances selected by $sel_r^n$. An example of $pattern_r^n$ is "the events $e_1, e_2, e_3$ have occurred in the order $e_1, e_2, e_3$, all three events are events regarding the same stock, and event $e_3$ occurred no later than 5 minutes after event $e_1$."

$eventType_r$ is the type of event inferred by this rule.

$mappingExpressions_r$ is a set of functions mapping the attribute values of the events that triggered this rule to the attribute values of the inferred event.

$prob_r \in [0, 1]$ is the probability of inferring the event by the rule. The exact semantics of this probability are defined in Section 4.2.

By definition, the predicates defined by $sel_r^n$ and $pattern_r^n$ are deterministic, as are the functions $mappingExpressions_r$. Therefore, the only uncertainty present in the rule is represented by the quantity $prob_r$. Indeed, many deterministic composite event languages can be viewed as defining a set of rules $R$ such that each rule $r \in R$ is of the form $\langle sel_r^n, pattern_r^n, eventType_r, mappingExpressions_r \rangle$.

We assume that an event type is either explicit or inferred by a single rule. This is because if there is more than one source of information for an event type (*e.g.*, two rules), the probabilities supplied by the separate sources (rules) must be reconciled to create a well-defined probability space.

The rule definition provides a *set* of languages $L$ for knowledge representation, by instantiating $sel_r^n$, $pattern_r^n$, $consumptionPredicates$, and $mappingExpressions_r$. We conclude this section with a definition of a language $l_1 \in L$. In this language, we have the following:

- $sel_r^n$ is of the form $\langle selExpression_1, \ldots, selExpression_n \rangle$, where $selExpression_i$ is a selection expression of the form $\varepsilon_i \in eventType_i$, with $eventType$ being a valid event type. Given an event history, $selExpression_i$ will select a single event, $\varepsilon_i$. The event $\varepsilon_i$ selected by $selExpression_i$ is the first event in the event history of type $eventType_i$ that was not selected by a selection expression $selExpression_j$ such that $j < i$.
- $pattern_r^n$ is a conjunctive predicate defined over the events $\varepsilon_1, \ldots, \varepsilon_n$ selected by $sel_r^n$, of the form $\wedge_{i=1}^m predicate_i$. $predicate_i$ is either a temporal predicate, or an equality relation between attributes. If $predicate_i$ is an equality predicate, it is of the form $\varepsilon_k.attribute_l = \varepsilon_j.attribute_m$ for $k \neq j$. This specifies that the value of $attribute_l$ of event $\varepsilon_k$ must be the same as $attribute_m$ of event $\varepsilon_j$. A temporal predicate $predicate_i$ takes one of the following forms:
  - $a \leq \varepsilon_k.occT \leq b$, where $a$ and $b$ are temporal constants denoting time points in the range $[0, \infty]$. This predicate specifies that the event has occurred within the interval $[a, b]$.
  - $\varepsilon_j.occT < \varepsilon_k.occT$ for $k \neq j$. This predicate defines a partial order over subsets of events.
  - $\varepsilon_j.occT \leq \varepsilon_k.occT \leq \varepsilon_j.occT + c$ for $k \neq j$, where $c$ is a temporal constant such that $c > 0$. This predicate specifies that an event has happened within a specified interval relative to another event.
- Regarding $mappingExpression_r$ the occurrence time of the inferred event is always determined to be the point in time at which the inference was carried out. As for other attributes, two types of functions are allowed. The first is a function mapping a specific attribute value of a specific event participating in $pattern_r$ to an attribute of the inferred event. The second is the mapping of a constant value to an inferred attribute.

We conclude this section with an example of a rule in $l_1$ and its partial application. Let the rule $r_1$ be a rule designed to recognize an illegal stock trading operation, defined as follows: $sel_{r_1}^n$ is $\langle \varepsilon_1 \in stockSell, \varepsilon_2 \in stockPurchase \rangle$, $pattern_{r_1}^n$ is $(\varepsilon_1.occT \leq \varepsilon_2.occT \leq \varepsilon_1.occT + 5) \wedge (\varepsilon_1.stockTicker = \varepsilon_2.stockTicker) \wedge (\varepsilon_1.customerID = \varepsilon_2.customerID)$, $eventType_{r_1}$ is $illegalStockTrading$, and $mappingExpression_{r_1}$ consists of two functions: The first maps $\varepsilon_1.stockTicker$ to the $stockTicker$ attribute of the inferred event, and the second maps $\varepsilon_1.customerID$ to the $customerID$ attribute of the inferred event. Finally, $prob_r$ is 0.7. The intuition underlying such a definition is that the sale of a stock, followed closely by a purchase of the same stock, is an indication of suspicious activity. Consider now the event history $e_1, e_2, e_3$, where the events $e_1$, $e_2$ and $e_3$ have the following associated information: $e_1 \in stockSell, e_1.occT = 5, e_1.stockTicker =$ "IBM", $e_1.customerID =$ "C1", $e_2 \in stockQuote$, $e_2.occT = 7$ and $e_3 \in stockPurchase$, $e_3.occT = 9$, $e_3.stockTicker =$ "IBM" and $e_3.customerID =$ "C1". In this case, $e_1$ and $e_3$ will be selected by $sel_{r_1}^n$, and assigned to $\varepsilon_1$ and $\varepsilon_2$ in $pattern_{r_1}^n$ respectively, which will cause $pattern_{r_1}^n$ to evaluate to $true$. However, if $e_3.stockTicker =$ "MSFT" or $e_3.occT = 11$, $pattern_{r_1}^n$ would have evaluated to $false$. In addition, if $e_3 \in stockQuote$, $pattern_{r_1}^n$ would not have been evaluated at all, as exactly two events have to be selected by $\langle E_1 \in stockSell, E_2 \in stockPurchase \rangle$ to evaluate $pattern_{r_1}^n$.

---

[1] The above formalism can be easily extended to cases in which a rule, when triggerred on a specific event history, results in the inference of more than a single event.

### 4.1 THE PROBABILITY SPACE

Rule reasoning facilities need to be able to compute at any point in time $t$ the probability that an event $e$, with specific data, occurred at some time $t' \leq t$. In addition, the only evidence that can be taken into account is that which is known to the system at time $t$. Therefore, a (possibly different) probability space is defined for each $t$. An intuitively appealing way to define this probability space involves possible world semantics (see (Halpern, 1990)). Using such a definition, the probability space at time $t$ is a tuple $P_t = (W_t, F_t, \mu_t)$ such that:

- $W_t$ is a set of possible worlds, with each possible world corresponding to a specific event history that is considered possible at time $t$. An assumption that holds in all practical applications is that the number of events in each event history, as well as the overall number of events, is finite. This is because an actual system cannot consider an infinite number of events in a finite time period. Therefore, each possible world corresponds to an event history that is finite in size. In addition, we assume that the real world is one of the possible worlds.
- $F_t \subseteq 2^{|W_t|}$ is a $\sigma$-algebra over $W_t$.
- $\mu_t : F_t \to [0,1]$ is a probability measure over $F_t$.

We call the above representation of the probability space the *possible world representation*.

There is also a less intuitive, but more computationally useful, way to define the probability space. Let $E_1, E_2, \ldots$ be the set of EIDs representing the information about all events of interest. It is clear that each finite event history can be represented by a finite number of values $e_1, \ldots, e_n$, such that there exist a finite number of EIDs $E_1, \ldots, E_n$ where $e_i$ is a possible value of $E_i$. Therefore, each possible world $w_t \in W_t$ can be represented by such a finite number of values. In addition, as the overall number of events is finite, there is a finite number of events $E_1, \ldots, E_m$ such that $E_i$ could have occurred in some $w_t \in W_t$. Finally, if $|W_t|$ is finite, each $Ev_i$ can only have a finite number of associated values (one for each world in $W_t$) in which it appears. Note that in such a case, each possible $w_t$ can be represented by a finite number of values $Val_1, \ldots, Val_m$, where the value $Val_1, \ldots, Val_n$ for some $n \leq m$ is a set of values, each such set representing the values of one of the $n$ events that occurred in $w_t$, and $Val_{n+1}, \ldots Val_m$ are all $\{notOccurred\}$. From this it follows that if the probability space $P_t$ represents the knowledge of the composite event system at time $t$, this knowledge can be represented by a set of $m$ EIDs - $E_1, \ldots, E_m$.

Therefore, in the case where $|W_t|$ is finite, it is possible to define the probability space $P_t$ as $(\Omega_t, F_t, \mu_t')$ where:

- $\Omega_t = \{Val_1, \ldots, Val_m\}$ such that the tuple $Val_1, \ldots, Val_m$ is the set of values corresponding to an event history, where this event history is a possible world $w_t \in W_t$ as described above. Obviously, $|\Omega_t|$ is finite.
- $F_t = 2^{|\Omega_t|}$
- $\mu_t'(\{Val_1, \ldots, Val_m\}) = \mu_t(w_t)$ such that $w_t$ is the world represented by $\{Val_1, \ldots, Val_m\}$

This representation is termed the *EID representation*.

Conversely, note that each possible set of values of EIDs, $\{Val_1, \ldots, Val_m\}$ corresponds to some event history. Therefore, given an EID representation of $P_t$, where $|\Omega_t|$ is finite, it is obviously possible to create the corresponding finite-size possible worlds representation by defining a possible world $w_t \in W_t$ for each distinct set of values $\{Val_1, \ldots, Val_m\}$.

### 4.2 RULE SEMANTICS

Based on the above probability space, we now define the semantics of rules. Intuitively, the semantics of each rule $r$ are as follows: Let $EH_{t_1}^{t_2}$ be an event history. If the rule $r$ is applied at some time $t \geq t_2$, and the set of events selected by $sel_r^n$ from $EH_{t_1}^{t_2}$ is of size $n$ and is such that $pattern_r^n$ on this event is true, then the event inferred by rule $r$ occurred with probability $prob_r$. In addition, in such a case, the value of its corresponding attributes is the value defined by $mappingExpressions_r$. Otherwise, the event cannot be inferred.

Formally, let $sel_r^n(EH_{t_1}^{t_2})$ denote the set of events selected by $sel_r^n$ from $EH_{t_1}^{t_2}$, and let $pattern_r^n(sel_r^n(EH_{t_1}^{t_2}))$ denote the value of the predicate $pattern_r^n$ on $sel_r^n(EH_{t_1}^{t_2})$ (recall that if $|sel_r^n(EH_{t_1}^{t_2})| < n$ then the rule is not applied). In addition, let $val_1, \ldots, val_n$ denote the value of the attributes of the inferred event $e_r$ as defined by $mappingExpressions_r$. Then, if the specific event history is known, and denoting by $E_r$ the EID corresponding to $e_r$, we have the following:

$$\Pr(E_r = \{occurred, val_1, \ldots, val_n\} | EH_{t_1}^{t_2}) = prob_r$$
$$\text{if } pattern_r^n(SEL_r^n(EH_{t_1}^{t_2})) = true \quad (1)$$

$$\Pr(E_r = \{notOccurred\} | EH_{t_1}^{t_2}) = (1 - prob_r)$$
$$\text{if } pattern_r^n(SEL_r^n(EH_{t_1}^{t_2})) = true \quad (2)$$

$$\Pr(E_r = \{notOccurred\} | EH_{t_1}^{t_2}) = 1$$
$$\text{if } pattern_r^n(SEL_r^n(EH_{t_1}^{t_2})) = false \quad (3)$$

Recall from Section 4.1 that if $|W_t|$ is finite, the probability space can be represented by a finite set of finite value RVs. In addition, note that the rule semantics defined above specify that the probability of the inferred event does not depend on the entire event history, but rather on the events selected by $sel_r^n$. Therefore, let us denote by $E_1, \ldots, E_m$ the set of EIDs that describe knowledge regarding the event history, and let $\{E_{i_1}, \ldots E_{i_l}\}$ describe the subset of $\{E_1, \ldots, E_m\}$ that are candidates for selection by $sel_r^n$ (note that $l \geq n$, as $sel_r^n$ must choose the first $n$ events that have **actually occurred**). An EID $E$ is a candidate for selection if there is a possible event history in the probability space $P_t$ such that there is a set of $n$ events which will be chosen by $sel_r^n$ from this event history, and the event $e$ corresponding to $E$ is in this set. Then for all sets of values $\{Val_1, \ldots, Val_m\}$ such that $E_i = Val_i$, we have that

$$\Pr(E_r | E_{i_1}, \ldots E_{i_l}) = \Pr(E_r | E_1, \ldots, E_m) \quad (4)$$

*i.e.*, $E_r$ is conditionally independent of $\{E_1, \ldots, E_m\} \setminus \{E_{i_1}, \ldots E_{i_l}\}$ given $\{E_{i_1}, \ldots E_{i_l}\}$. Now let $Val_{i_1}, \ldots, Val_{i_m}$ denote a specific set of values of $E_{i_1}, \ldots E_{i_l}$. Given such a set of specific values, the subset $\{e'_{j_1}, \ldots, e'_{j_n}\}$ selected by $sel_r^n$ is well defined. Therefore, we have from the above equations that:

$$\Pr(E_r = \{occurred, val_1, \ldots, val_n\} | Val_{i_1}, \ldots, Val_{i_m})$$
$$= prob_r \text{ if } pattern_r^n(e'_{j_1}, \ldots, e'_{j_n}) = true \quad (5)$$

$$\Pr(E_r = \{notOccurred\} | Val_{i_1}, \ldots, Val_{i_m}) = (1 - prob_r)$$
$$\text{if } pattern_r^n(e'_{j_1}, \ldots, e'_{j_n}) = true \quad (6)$$

$$\Pr(E_r = \{notOccurred\} | e_{i_1}, \ldots, e_{i_m}) = 1$$
$$\text{if } pattern_r^n(e'_{j_1}, \ldots, e'_{j_n}) = false \quad (7)$$

## 5 PROBABILITY CALCULATION

The general principles underlying the method for calculating probabilities is the use of the KBMC paradigm. The main reason for using this paradigm is that the underlying probability space changes over time as new evidence regarding events reaches the system. Therefore, different models for calculating probabilities should be created for different time points.

While the type of model depends on the specific language for which inference is carried out, we believe that there are many cases in which a simple Bayesian network can be used. For such cases the following assumptions should hold:

- $P_t$ is such that $|W_t|$ is finite.
- For each event, there is no uncertainty regarding the time of its occurrence, *i.e.*, there is a specific $t$ such that either the event occurred at $t$, or did not occur at all.
- There is a mechanism in the language or the execution model for guaranteeing *termination*. This means that for each event history $EH_{t_1}^{t_2}$, only a finite number of rules will be triggered (this ensures that only a finite number of EIDs can be added to the event history). One way to ensure termination is to avoid cycles in rule definition (see (Patton, 1998)).
- There is a mechanism for guaranteeing *determinism*. Determinism guarantees that the application of the same set of rules to a given event history will always result in the same probability space. Defining a full order on the application of rules by assigning each rule a priority is one way to guarantee determinism (see (Patton, 1998)).

An algorithm for constructing a Bayesian network for the general case, under these assumptions, is not described in this paper. Rather, we will detail the algorithm for generating the Bayesian network for the language $l_1$ defined in Section 4.2. However, before describing the algorithm, we would like to stress two main features that distinguish our implementation of the KBMC paradigm from many existing applications of this paradigm. First, the Bayesian network is dynamically updated as information about events reaches the system. This is to ensure that the constructed network reflects, at each time point $t$, the probability space $P_t$. Second, throughout the inference process we maintain additional information beyond the Bayesian network. This additional information is used both to allow an efficient dynamic update of the network, and to make the inference process more efficient.

In order to define the construction algorithm, we first assume that the set of rules has no cycles, and that a priority is assigned to each rule so that determinism is guaranteed. Now recall that by definition, the occurrence time of each inferred event in $l_1$ is the time in which the rule was applied. Therefore, the single possible occurrence time of each EID defines a full order on the EIDs (we will denote this single point in time for EID $E$ by $E.occT$). In addition, according to Eq. 4, each EID is independent of all preceding EIDs, given the EIDs that may be selected by the selection expression. Therefore, we construct a Bayesian network such that the nodes of the network consist of the set of random variables in the system event history, and an edge exists between EID $E_1$ and EID $E_2$ iff $E_1.occT \leq E_2.occT$ and $E_2$ is an EID corresponding to an event that may be inferred by rule $r$, where the event corresponding to $E_1$ may be selected by $sel_r^n$. A network constructed by these principles encodes the probabilistic independence required by Eq. 4 (see (Pearl, 1988)). This structure is now augmented with values based on Eq. 5-7.

Based on the above principles, a Bayesian network is constructed and dynamically updated as events enter the system. At each point in time, nodes and edges may be added to the Bayesian network. The algorithm below describes this dynamic construction. The information regarding the new event is represented by some EID $E$, the system event history is represented by $EH$, and the constructed Bayesian network by $BN$. The algorithm follows:

1. $EH \leftarrow EH \cup \{E\}$
2. Add a node for $E$ to $BN$.
3. For each such rule $r$ in a decreasing priority order:
   (a) Denote by $sel_r^n(EH)$ the subset of EIDs in $EH$ that may be selected by $sel_r^n$ (these are all EIDs whose *type* attribute is of one of the types specified by $sel_r^n$).
   (b) If there is a subset of events in $sel_r^n(EH)$ that may be selected by $sel_r^n$ such that $pattern_r^n$ is true, add a vertex for the inferred event's EID $E_r$. In addition, add edges from all events in $sel_r^n(EH)$ to the event $E_r$.
   (c) For $E_r$, fill in the quantities for the conditional probabilities according to Eq. 5-7.
4. Calculate the required occurrence probabilities in the probability space defined by the constructed Bayesian network.

The above algorithm describes at a high level the calculation of the required probabilities, omitting the details of several steps. The omitted details include the mechanism for selection of events as indicated by $sel_r^n$, the evaluation of the predicates defined by $pattern_r^n$, and the exact algorithm used to infer the required probabilities from the Bayesian network. In all of these cases, standard algorithms from the domains of deterministic event composition and Bayesian networks may be used and extended. The specific algorithms used for these tasks will determine the complexity of our algorithm. However, the dominant factor will be the calculation of the required probabilities from the Bayesian network, which is known to be computationally expensive. Therefore, ways to speed up this step is a topic that warrants future research (see discussion in Section 6).

### 5.1 INFERENCE EXAMPLE

This section illustrates the above algorithm for language $l_1$ using a specific example. Assume that in the system there exists the rule $r_1$, as defined in Section 4. Now let us assume that information is received about the possible occurrence of an event $e_1$ such that $e_1 \in stockSell, e_1.occT = 5, e_1.stockTicker = "IBM", e_1.customerID = "C1"$. In addition, the probability that this event occurred is 0.6. This information is represented in the system by an EID $E_1$ with two possible states: $\{notOccurred\}$ and $\{occurred, stockSell, 5, IBM, Customer1\}$ (we will abbreviate the second state by $\{occurred\}$. The constructed Bayesian network will consist of a single node $E_1$ with $\Pr(E_1 = \{notOccurred\}) = 0.4$ and $\Pr(E_1 = \{occurred\}) = 0.6$. Now assume that information is received regarding the possible occurrence of an additional event $e_2 \in stockSell, e_2.occT = 9, e_2.stockTicker = "IBM", e_2.customerID = "C1"$. This is represented in the system by the EID $E_2$ with two states as above, which is added to the Bayesian network. At this stage, the Bayesian network consists of two disconnected nodes, as depicted in Figure 1(a).

Note that although two possible events have occurred, there is no possible world in which two events are selected by $sel_{r_1}^n$, and, therefore, $r_1$ is not triggered. Now, assume that information regarding a third event $e_3$ reaches the system, such that $e_3 \in stockPurchase, e_3.occT = 5, e_3.stockTicker = "IBM", e_3.customerID = "C1"$. This is represented in the system by the EID $E_3$. Now there is one possible world in which there is a non-zero probability that $E_r$ occurs - this is the world in which the event history is $e_2, e_3$. Therefore, a node $E_r$ is added to the network, and edges will be added from $E_1, E_2, E_3$ to $E_r$. This will result in the Bayesian network depicted in Figure 1(b).

In addition, the event corresponding to the EID $E_r$ occurs only if $e_1$ did not occur, and $e_2$ and $e_3$ both occurred. Therefore, according to Eq. 5-7, $\Pr(E_r = \{occurred\}|E_1 = \{notOccurred\}, E_2 = \{occurred\}, E_3 = \{occurred\}) = 0.7$, and $\Pr(E_r = \{occurred\}|E_1, E_2, E_3) = 0$ for all other value combinations of $E_1$, $E_2$ and $E_3$.

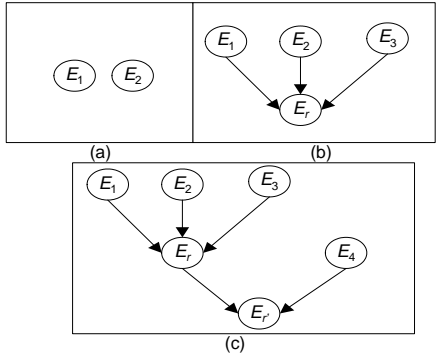

Figure 1: Constructed Bayesian Networks

Finally, if we define an additional rule $r'$ which states that an event has a non-zero occurrence probability whenever $e_r$ and an additional event of type $e_4$ occurs, and $e_4$ is signaled, this will result in the network depicted in Figure 1(c)

## 6 DISCUSSION

In this paper, we propose a model for defining composite event systems that takes into account the uncertainty inherent in many active systems such as BAM and BPM systems. The formal framework was introduced, enabling both knowledge representation and probability inference. The representation of uncertainty in such systems, as well as the probability space, were defined. In addition, the principles enabling the calculation of the probabilities were introduced, providing a formal framework for handling several types of uncertainty inherent in many active applications. To the best of our knowledge, this is the first formal and comprehensive treatment of uncertainty in event composition languages.

The representations of uncertainty, defined in sections 3.2 and 4, together with inference algorithms of the type outlined in Section 5, can serve as the basis of a component that can be embedded in any active system. Such a component would be used as follows: During the development of the active application, the application-specific uncertainty will be specified. An example of such specification is the rule defined in Section 4. Based on these definitions, as events are signalled during the execution of the active system, the probabilities of the inferred events can be constantly calculated and updated. Therefore, such a component would enable any active system to handle uncertainty of the type discussed in this paper in a general manner. Additionally, such a component will have strong formal underpinnings, as defined in this article.

There are several possible future avenues for research. The model of composite event systems can be extended to enable predictions regarding the occurrence of future events. Although the rule representation makes it possible to infer that events will occur in the future, other extensions are required to handle prediction – for instance, a mechanism that enables specifying the probabilities of future events being signalled by outside sources. Additional future work lies in improving the efficiency of the inference algorithms described in Section 5. One possible manner in which improvements can be achieved is the incorporation of techniques such as the one in (Flores et al., 2003) for incremental updating of the junction tree used for probability calculation. Such incremental updating would significantly reduce the effort required for re-creation of the junction tree after each update to the Bayesian network.